\documentclass[runningheads]{llncs}
\usepackage[T1]{fontenc}
\usepackage{graphicx}
\usepackage{amsmath}
\usepackage{booktabs}
\usepackage{multirow}
\usepackage{amsfonts}
\begin{document}
\title{Unified MRI Brain Image Translation via
Hierarchical Tumor Structure Comparison}
\titlerunning{Unified MRI Brain Image Translation via HTSCGAN}
%
%
\author{Yupeng Cai\inst{1} \and
Jia Wei\inst{2}\thanks{Corresponding author}\and
Jianlong Zhou\inst{3}}
\authorrunning{Y. Cai et al.}
%
\institute{South China University of Technology, Guangzhou 510640, China \email{202321044601@mail.scut.edu.cn} \and
South China University of Technology, Guangzhou 510640, China
\email{csjwei@scut.edu.cn}\\
\and
UTS Data Science Institute, University of Technology Sydney, Ultimo, NSW 2007, Australia\\
\email{jianlong.zhou@uts.edu.au}}
%
\maketitle              
\begin{abstract}
Multi-modal MRI brain image translation via available modalities holds significant practical importance in modern medicine, providing robust support for early diagnosis, treatment planning, and outcome assessment of diseases. For this purpose, it is important to ensure the fidelity of the tumor regions after translation. However, existing brain image translation methods ignore the structure information of different tumor regions, which could assist translation models in enhancing the quality and clinical applicability of the translated images. 
 In this work, we propose a novel translation model called HTSCGAN, which is a unified multi-modal brain image translation generative adversarial model integrating the structural information within tumor regions with the aim of improving the quality of brain image translation. Specifically, the generator employs three Patch Contrast Module (PCM) with different patch sizes to capture the hierarchical structural information of the tumor regions. In addition, a pretrained Patch Classifier (PC) and a pretrained Structure-Aware Encoder (SAE) are employed to derive the generated image containing the same tumor region structure as the ground truth image via patch classification loss and tumor perceptual loss, respectively. The experiments on BraTS2020 and BraTS2021 demonstrate strong performance of our model in both translation tasks and down stream segmentation tasks, highlighting its effectiveness in enhancing the quality and clinical relevance of the translated brain images. Our code is available at \url{https://anonymous.4open.science/r/HTSCGAN}.

\end{abstract}
\section{Introduction}

In the field of medical image analysis, the acquisition of multi-modal medical images (such as CT, MRI, and PET) plays a critical role in clinical diagnosis and medical research \cite{wu2010multiparametric,dickinson2013clinical}. However, in clinical practice, obtaining a complete set of multi-modal images for each patient is not always be feasible \cite{havaei2016hemis}, and the technical and equipment differences across these modalities result in significant variability among the images, presenting significant challenges for comprehensive diagnosis. To overcome this issue, there is a pressing need for innovative methods to translate existing medical images into additional modalities with high quality and fidelity, thereby improving the understanding of pathological conditions and enhancing diagnostic accuracy \cite{xie2021cotr}.

In recent years, numerous models for medical image translation are proposed, significantly improving the efficiency of medical diagnosis tasks. For instance, GANs have been applied to medical image translation \cite{goodfellow2020generative,armanious2020medgan}, as well as diffusion models \cite{ozbey2023unsupervised,zhan2024medm2g} and Transformer frameworks \cite{dosovitskiy2020image,li2020tumorgan}. However, most of these models overlook the critical structural information of tumor regions in medical image translation. Accurate translation of tumor regions is essential for medical diagnosis, as erroneous representations can mislead diagnostic conclusions and adversely affect subsequent tasks. While some models, such as SPMI\cite{kang2023structure} and RegistFormer\cite{kim2025improving}, do incorporate structural information, their approaches encounter difficulties in adapting to brain imaging. This is largely due to the complex structural relationships inherent in brain images and the significant differences in modalities present within the tumor regions.

Generating realistic tumor regions poses several challenges: 1) Tumor structures are highly complex, with typical brain tumors comprising three distinct regions: non-enhancing tumor (NT), enhancing tumor (ET), and edema (Ed). These regions exhibit intricate structural interrelationships. 2) Tumor regions occupy a relatively small portion of the overall medical image, which increases the likelihood that models may either ignore these areas or pay insufficient attention to the crucial structural information of the tumor regions \cite{adam2020grainger}. 3) Tumor-related information differs across imaging modalities. For example, the Flair and T2 modalities emphasize the visualization of the entire tumor area, whereas T1ce is primarily used to differentiate tumor tissue from non-tumorous lesions, and T1 provides detailed anatomical structure \cite{jog2017random}. This variability in modality-specific information presents significant challenges for achieving effective modality translation. To address these challenges, we propose a novel model named Hierarchical Tumor Structure Comparison GAN (HTSCGAN), as illustrated in Figure 1. HTSCGAN is a unified generative adversarial model for medical image translation that supports arbitrary modality inputs and outputs. In the model, we incorporate the structural information within tumor regions into the image translation process, enhancing the model's ability to generate more accurate and clinically meaningful images. We achieve this through three key perspectives as follows.

As discussed in \cite{gong2021vision}, feature similarity between different patches of the same image tends to increase as the network depth grows. Given the small proportion of the tumor regions, this often leads to the tumor's complex structural information being attenuated during training. To address this problem, we introduce Patch Contrast Module (PCM), a method based on contrastive learning between patches within the image, which ensures the independence of tumor-specific features and enhances the diversity and distinctiveness of the learned representations (Section 3.2).

The significant variation in tumor characteristics across different modalities presents challenges in capturing the tumor features. To accurately represent the tumor features of the target modality, we propose Patch-Classifier (PC). Each modality is paired with a specific pretrained PC designed to distinguish between different tumor regions based on the corresponding patches. Furthermore, inspired by knowledge distillation, we propose a training strategy where the tumor regions predictions of the generated images are encouraged to closely match those of the real images, as predicted by the classifier, rather than directly aligning with the ground truth soft labels. This strategy effectively prevents the generator from learning inaccurate tumor features (Section 3.3).

Additionally, we acknowledge that various structural regions are interrelated, emphasizing the importance of capturing these dependencies in our model. For instance, the enhancing tumor (ET) often forms a ring around the non-enhancing tumor (NT), while edema (Ed) tends to surround both regions and display more irregular shapes, resulting in a concentric or nested appearance. To capture these structural dependencies, we introduce a pre-trained structure-aware encoder that learns these relationships by reconstructing images with partially missing tumor regions. (Section 3.4).

We evaluate the performance of HTSCGAN through extensive experiments. Our main contributions can be summarized as follows:

•	We propose a method to extract structural information from tumor regions and integrate it into the translation task to enhance model performance.

•	We propose the PCM, which applies contrastive learning across patches to  capture finer-grid features in small, distinct regions.

•	Our model demonstrates superior performance in both medical image translation and downstream segmentation tasks, validating its utility and effectiveness.

\section{Related Works} 
\label{sec:formatting}

\subsection{Transformer}

Transformer capture long-range dependencies by computing attention between tokens, thereby enhancing the model’s representational power. Vision Transformer (ViT) \cite{dosovitskiy2020image} was the first to apply Transformers to vision tasks by dividing images into patches and computing global attention across them. Building on this, Swin-Transformer \cite{liu2021swin} improved efficiency by introducing window-based attention and a sliding window mechanism. In recent years, there has been a surge in applying Transformers to medical image translation tasks. For example, CyTran \cite{ristea2023cytran} incorporates Transformers into GANs while also introducing cycle-consistency loss, and ResViT \cite{dalmaz2022resvit} skillfully combines CNNs with Transformers, proposing a unified model for image translation. MMTransformer \cite{liu2023one} further introduces a multi-contrast, multi-scale Transformer module that leverages a Swin-like window mechanism, effectively capturing intra- and inter-modal relationships.

\subsection{Contrastive Learning}

Contrastive learning is an unsupervised or self-supervised learning method aimed at maximizing the similarity between positive sample pairs while minimizing the similarity between negative pairs. FaceNet \cite{schroff2015facenet} first introduced the triplet loss for contrastive learning, optimizing the model by adjusting the distance between positive and negative samples. Several studies have incorporated patch-level contrastive loss into image generation frameworks, exemplified by approaches such as AMUIT\cite{liu2023contrastive}. In these methods, the contrastive learning objective typically designates patches from identical spatial locations in the original and generated images as positive pairs, while patches from differing locations are treated as negative pairs. However, this strategy appears ineffective for brain image generation. This ineffectiveness arises because tumor features across different modalities can exhibit substantial discrepancies even when located at the same spatial position. In contrast, within a single modality, features corresponding to the same tumor region, albeit at varying spatial locations, tend to be similar. Consequently, the conventional pairing of positive and negative samples in contrastive learning does not yield the desired efficacy in the context of brain image generation.

\subsection{Medical image translation}

In recent years, numerous advanced machine learning approaches have been applied to medical image translation tasks, resulting in a surge of noteworthy contributions. Medical image translation models can generally be classified into two categories. The first category includes models that support specific single- or dual-modality conversion. For instance, SA-GAN \cite{yu2020sample} introduces a sample-adaptive pathway and employs a self-attention mechanism to capture long-range dependencies in images, while SynDiff \cite{ozbey2023unsupervised} uses a conditional diffusion model for image translation.The second category encompasses models that unified multi-modal medical image translation. For example, MILR \cite{chartsias2017multimodal} uses a multi-encoder architecture to map different modalities into a shared latent space for integration and then generates the target modality using separate decoders. Similarly, MMGAN \cite{sharma2019missing} combines U-Net with GAN and proposes a padding-based approach to address channel mismatch by filling in missing modalities. TSF \cite{han2023explainable} formulates modality translation as a sequence-to-sequence prediction task. However, these models do not account for the structural relationships within tumor regions, resulting in suboptimal synthesis outcomes in these areas. Some studies, such as SPIT
, have incorporated structural relationship information of ocular blood vessels, while RegistFormer\cite{kim2025improving} has also considered the structural information of the pelvic region. Nevertheless, these approaches are not transferable to brain imaging, as brain tumors do not represent a singular structure; instead, they encompass complex structural relationships and exhibit significant differences among modalities within the tumor regions.

\section{Method}

\begin{figure}
  \centering
  \includegraphics[width=0.9\linewidth]{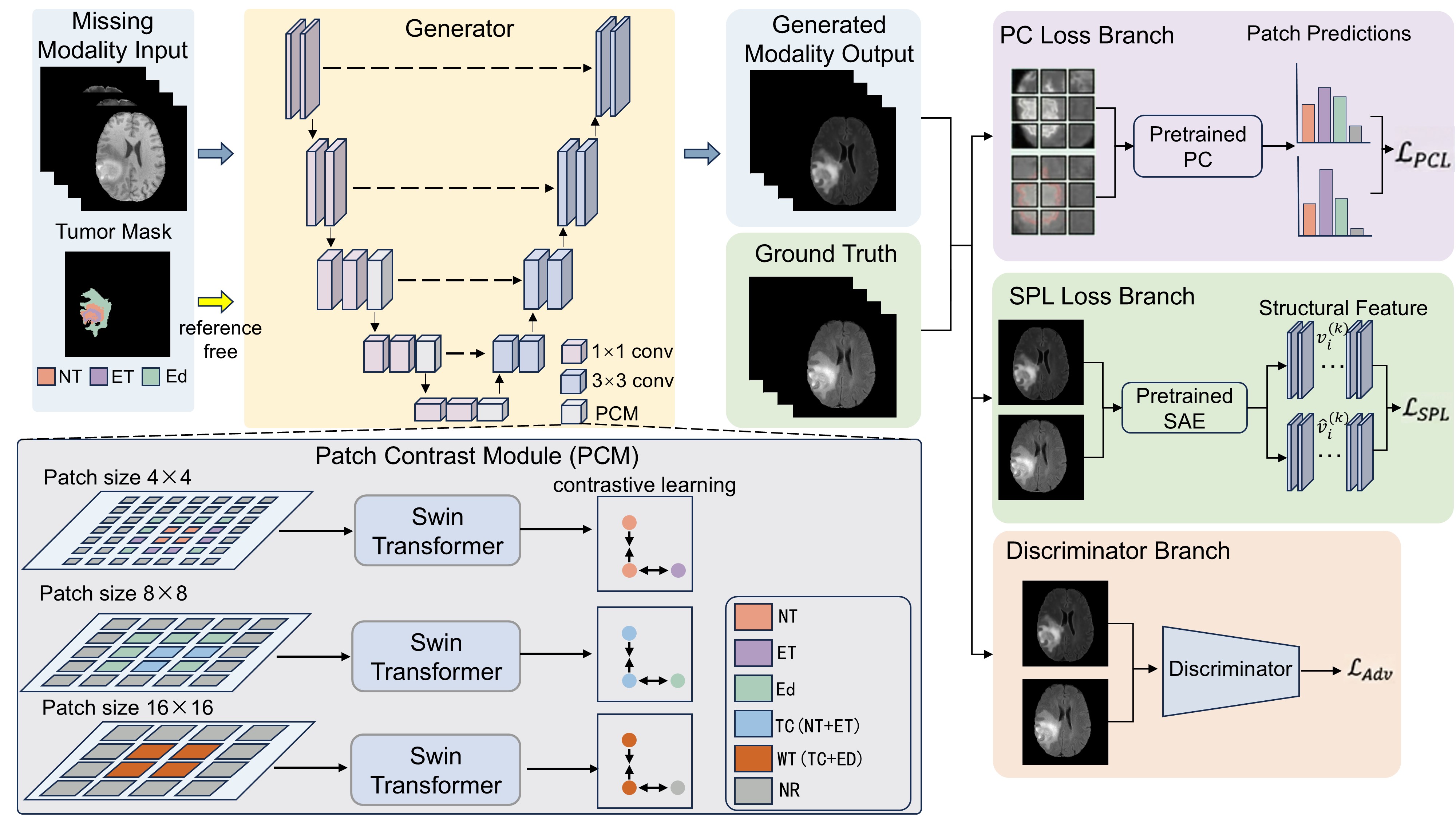}
  \caption{Overview of HTSCGAN.}
  \label{fig:1}
\end{figure}

The architecture of HTSCGAN is illustrated in Figure 1, which consists of a generator with three PCMs, a pre-trained SAE and a pre-trained PC. In the training phase of the model, tumor segmentation labels are utilized; however, during the inference phase, labels are not required.

\subsection{Generator}
Equipped with a U-Net \cite{ronneberger2015u} architecture, our generator incorporates transformer modules for contrastive learning.
In the generator, 1×1 convolutions and Swin-Transformer are used for the encoder along with downsampling to replace the patch extraction operation, while 3×3 convolutions are applied for the decoder. This design enhance the contrastive learning performance by effectively extracting more distinct features. The Transformer modules expand the receptive field, enabling the model to focus on both local and global features. To achieve consistent brain image translation, we address variations in input modalities by applying zero-padding, where each input modality corresponds to a specific channel. Generator training is guided by a combination of contrastive loss (from contrastive learning), patch classification loss (from a pre-trained PC), structure perceptual loss (from a pre-trained SAE), adversarial loss (from discriminator), and translation loss, as detailed in Section 3.6. These constraints jointly reinforce image translation fidelity and feature alignment.

\subsection{Patch Contrast Module}\label{sec:contrast}
As illustrated in Figure 1, PCM utilizes Swin-Transformer \cite{liu2021swin} architecture, it maintains the distinctiveness of features across different tumor regions through contrastive learning between image patches. The generator applies PCM at three scales, with patch sizes of 4×4, 8×8, and 16×16. The region division changes with different scales: for patch size 4×4, the image is divided into the four standard regions: NT, ET, Ed and normal region (NR); for patch size 8×8, the NT and ET are merged into tumor core (TC); and for patch size 16×16, the image is split into tumor and NR. This multi-scale design enhances feature representation by capturing information at different scales, thereby improving the model's robustness and overall performance.

During training, we randomly select a patch as the anchor, with patches from the same region serving as positive samples and patches from different regions acting as negative samples, and then leverage the features generated from the PCM to compute the contrastive loss for the optimization of the generator. Considering the different sizes of the regions, we choose neighboring regions as negative samples to better distinguish features from nearby areas. In each image, 50 patch pairs are chosen, and only patches covering more than 75\% of the largest region are selected. This strategy minimizes the influence of overlapping regions during training.
\begin{figure}[t]
  \centering
  \includegraphics[width=0.9\linewidth]{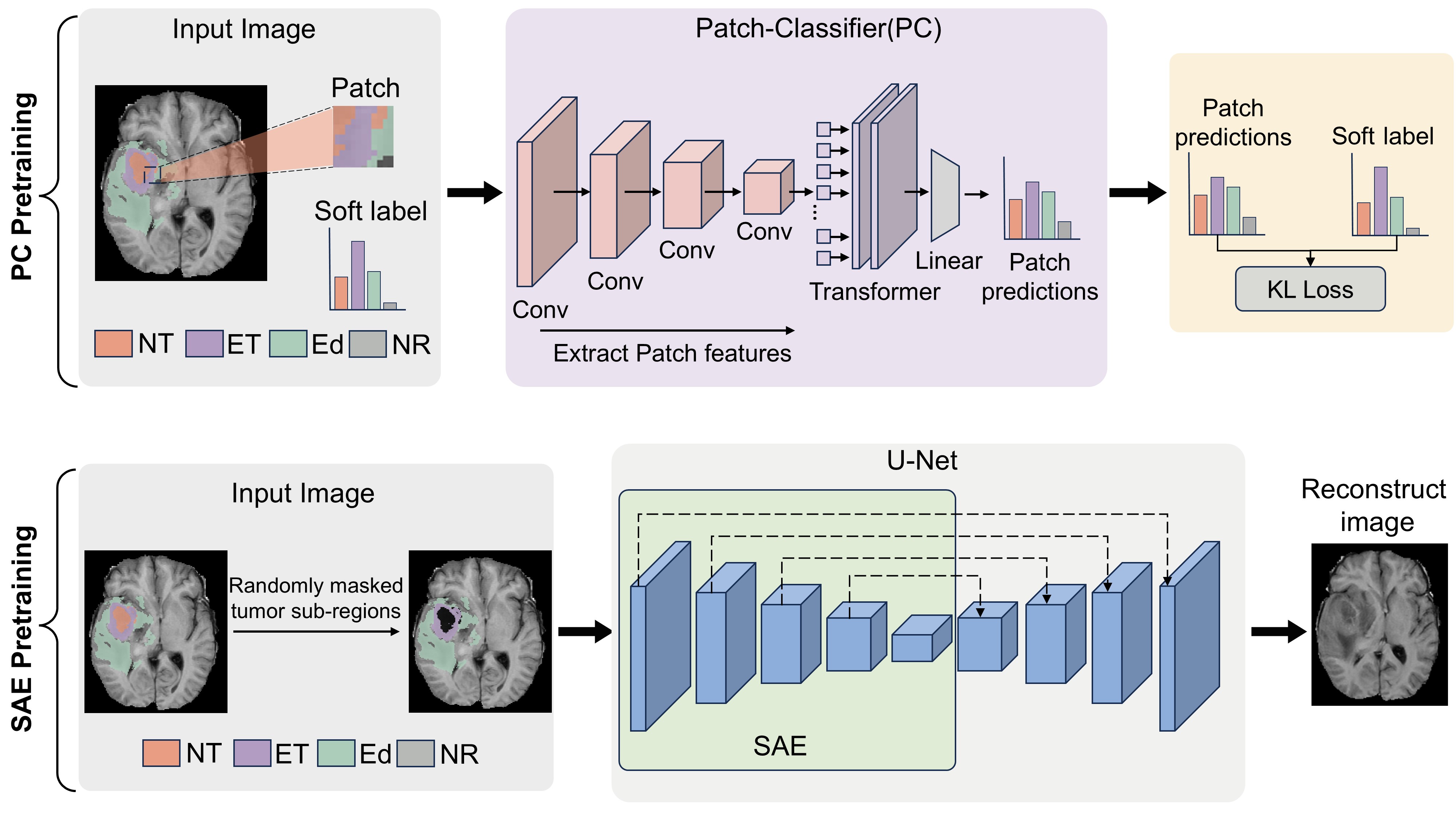}
  \caption{ Pretraining Process for Patch-Classifier (PC) and Structure-Aware Encoder (SAE).}
  \label{fig:3}
\end{figure}

\subsection{Patch-Classifier}\label{sec:classifier}
In our model, PC is used to categorize brain image patches and guide the model in learning accurate tumor region features, ensuring that the contrastive learning-derived features reflect the true characteristics of the tumor. Patches are divided into four categories (ET, NT, Ed, NR), and soft labels are assigned based on the region proportions within the patch, which can be formulated as follows:
\begin{equation}
    \text{Soft\_Label}_{i} = \frac{\text{NoP}(R_i)}{\text{NoP}(R)}
\end{equation}
Where \( i \in \{\text{NT, ET, Ed, NR}\} \), \text{NoP}() denotes the Number of Pixels in a specific area, $R$ represents the total area within the patch, and $R_i$ denotes the area of category \( i \) within the patch.

As shown in Figure 2, the classifier architecture integrates both CNN and Transformer components, leveraging the Transformer’s strengths in classification tasks. During the pretraining phase, the classifier focuses on classifying patches of real images, aiming for the results that align closely with the soft labels. This process resembles knowledge distillation, in terms of the model learns from the soft labels generated by a more powerful teacher model, effectively capturing essential features for classification. In the generator training phase, the parameters of the PC will be frozen. Rather than directly aligning with soft labels—which may lead to some tumor features being overly emphasized in the target modality, where such features are typically less pronounced—the objective is to align the classification results of the generated images with those of the real images. This strategy thus focuses the model's attention on tumor-specific features for the target modality, enhancing brain image translation performance by aligning with real data characteristics.  

\subsection{Structure-aware Encoder}\label{sec:structure}
In our model, SAE employs the U-Net architecture for brain image reconstruction, which is designed to model the relationships between tumor regions. As shown in the Figure 2, inspired by MAE \cite{he2022masked}, our pre-training approach for the encoder involves masking one or more randomly selected tumor regions while leaving at least one region unmasked. The model is then tasked with reconstructing the masked areas, thereby facilitating the learning of structural features of tumor regions.

In the generator training phase, the generated images and their tumor regions are passed through the SAE, whose parameters remain frozen during this phase. The SAE aligns its resulting features with those from the ground truth images and their tumor regions, ensuring consistency in tumor region structure feature representation.

\subsection{Loss Function} \label{sec:loss}
\noindent\textbf{Contrastive Loss.} The contrastive loss maximizes the distance between the anchor and positive samples, while pushing the anchor farther from negative samples. The loss function is based on InfoNCE \cite{he2020momentum} loss, known for its robustness to noise, scalability with large datasets, and simplicity, making it effective for contrastive learning tasks, which can be defined as follows:

\begin{scriptsize} 
\begin{equation}
    \mathcal{L}_{I} = - \frac{1}{M} \sum_{i=1}^{M} \frac{1}{N} \sum_{j=1}^{N} \log \frac{\exp(\mathbf{z}_{i,j} \cdot \mathbf{z}_{i,j}^+ / \tau)}{\exp(\mathbf{z}_{i,j} \cdot \mathbf{z}_{i,j}^+ / \tau) +  \exp(\mathbf{z}_{i,j} \cdot \mathbf{z}_{i,j}^- / \tau)} \\
\end{equation}
\end{scriptsize}Where $z_{i,j}$ is the $j$-th selected anchor patch in the contrastive learning module of layer $i$, while $z_{i,j}^{+}$ and $z_{i,j}^{-}$ denote its corresponding positive and negative patch pairs. The temperature parameter $\tau$ controls the separation between patches, $N$ denotes the number of anchor patches selected, and $M$ represents the number of layers used for contrastive learning, where we set $M$=3 and $N$=50 in our experiments.

\noindent\textbf{Patch Classification Loss.} Instead of hard labels, we assign soft labels based on the proportion of each region within a patch. During the pretraining phase, the labels serve as the "teacher," similar to traditional knowledge distillation, while the classifier's outputs serve as the "student." The KL divergence is employed to minimize the discrepancy between the two outputs.
\begin{equation}
    \mathcal{L}_{PCL}(P || Q) = \frac{1}{C} \sum_{i=1}^{C}\sum_{j} P(p_{i,j}) \log \frac{P(p_{i,j})}{Q(p_{i,j})}
\end{equation}
Where $p_{i,j}$ is the $j$-th patch of the $i$-th modality that needs to be translated, $P(p_{i,j})$ is the prediction for the $j$-th patch of the generated image from the patch-classifier, $Q(p_{i,j})$ is the prediction for the $j$-th patch of the real image from the patch-classifier, and $C$ is the number of modalities needs to be translated.

During the generator training phase, the classification results from the ground truth images serve as the "teacher," while the outputs from the classifier on the generated images act as the "student." The objective is to minimize the difference between these outputs, ensuring that the generated images closely resemble the ground truth in terms of classifier predictions.

\noindent\textbf{Structure Perceptual Loss.} Similar to perceptual loss \cite{johnson2016perceptual}, we utilize an structure-aware encoder to compute the structural dependency differences between the generated images and the ground truth, with a specific focus on the tumor regions. Features are extracted from the five layers of the SAE and apply L2 loss at each layer. This approach allows for a more comprehensive consideration of both global and local features. The loss is formally defined as:
\begin{equation}
    \mathcal{L}_{SPL}  = \frac{1}{C} \sum_{i=1}^{C} \frac{1}{L} \sum_{k=1}^{L} (v_i^{(k)} - \hat{v}_i^{(k)})^2
\end{equation}
Here, $v_i^{(k)}$ represents the feature vector of the generated image at the $k$-th layer of the encoder, while $\hat{v}_i^{(k)}$ denotes the feature vector of the real image at the $k$-th layer of the encoder, $L$ denotes the number of layers of features obtained through SAE, where $L$ is set to 5 in our experiments.

\noindent\textbf{Adversarial Loss.} The adversarial loss encourages the generator to produce more realistic images capable of fooling the discriminator. Specifically, we implement this loss using the L2 loss function to optimize the generator's ability to generate convincing images. This loss can be formulated as follows:

\begin{equation}
    \mathcal{L}_{Adv} = \frac{1}{C} \sum_{i=1}^{C} \mathbb{E} \left[ \left( D_{i}(y_{i}) - \text{Label}_{r} \right)^2 \right] .
\end{equation}
Where $D_{i}$ represents the discriminator for the corresponding generated modality, and $y_{i}$ represents the generated modality image, and $\text{Label}_{r}$ refers to the label of the real image.

\noindent\textbf{Translation Loss.} The translation loss is defined as the pixel-wise similarity between the generated images and the ground truth. This loss can be formulated as: 
\begin{equation}
    \mathcal{L}_{TL} = \frac{1}{C} \sum_{i=1}^{C} |y_i - \hat{y}_i|
\end{equation}
Where $y_i$ denotes the generated image, $\hat{y}_i$ represents the ground truth.

\noindent\textbf{Overall Loss.} The overall loss of the generator $\mathcal{L}_{G}$ can be defined as:
\begin{equation}
\begin{split}
    \mathcal{L}_{G} = & \; \lambda_1 \mathcal{L}_{\text{I}}  + \lambda_2 \mathcal{L}_{PCL} + \lambda_3 \mathcal{L}_{SPL} + \lambda_4 \mathcal{L}_{Adv}+\lambda_5 \mathcal{L}_{TL}
\end{split}
\end{equation}
In our experiments, we set $\lambda_1=1$, $\lambda_2=0.01$, $\lambda_3=0.1$, $\lambda_4=0.1$, $\lambda_5=10$.

\section{Datasets and Implementation Details}

\textbf{Dataset.} We trained our model using the BraTS2020 \cite{menze2014multimodal} and UCSF \cite{calabrese2022university} datasets. For both datasets, we performed center cropping to resize the images to 192×192 pixels, then apply bicubic interpolation to upscale them to 256×256. All images were normalized to the range [-1, 1]. To ensure that the brain structure and tumor region were adequately represented, we selected 20 slices per patient, which captured a substantial portion of both the brain and tumor regions.

\noindent\textbf{Implementation Details.} Our model was trained on an NVIDIA 3090 GPU, employing a random modality dropout strategy, similar to MMGAN \cite{sharma2019missing}, to simulate diverse input-output scenarios. Specifically, in the first 10 steps, only one modality was dropped; between steps 10 and 30, one or two modalities were randomly dropped; after 30 steps, any combination of missing modalities was allowed. The model was trained for 500 epochs, with the generator learning rate set to $2 \times 10^{-4}$ and the discriminator’s learning rate to $1 \times 10^{-4}$. 

For classifier training, a patch size of 16 was selected. The classifier was pre-trained for 100 epochs with a learning rate of $1 \times 10^{-4}$. For the structure-aware module, we pre-trained for 100 epochs with a learning rate of $2 \times 10^{-4}$. The Adam optimizer was employed for model training.

\begin{table*}[t]
    \centering
    \caption{Quantitative performance of the proposed HTSCGAN model compared with other methods on BraTS2020 and UCSF datasets. The data presented in the table represents averaged results across all input combinations, except for the comparison with SynDiff and RegistFormer, which only supports single-modality translation. The best performance is highlighted in bold, and statistical significance is confirmed by the Wilcoxon signed-rank test (p < 0.005).}\label{tab:1}
    \resizebox{1.0\textwidth}{!}{
        \begin{tabular}{c ccc ccc}
            \toprule
            \multirow{2}{*}{Method} & \multicolumn{3}{c}{BraTS2020} & \multicolumn{3}{c}{UCSF} \\ 
            \cmidrule(lr){2-4} \cmidrule(lr){5-7}
            & SSIM↑ & PSNR↑ & LPIPS↓ & SSIM↑ & PSNR↑ & LPIPS↓ \\ 
            \midrule
            MILR \cite{chartsias2017multimodal} & 0.8281±0.0406 & 23.46±3.02 & 0.1169±0.0331 & 0.8321±0.0412 & 23.51±3.14 & 0.1141±0.0323 \\
            TSF \cite{han2023explainable} & 0.8395±0.0414 & 23.96±2.94 & 0.1137±0.0311 & 0.8651±0.0437 & 25.09±3.31 & 0.0948±0.0282  \\
            MMGAN \cite{sharma2019missing} & 0.8341±0.0408 & 23.76±3.07 & 0.1220±0.0328 & 0.8560±0.0429 & 24.61±3.06 & 0.1098±0.0298 \\ 
            MMTransformer \cite{liu2023one} & 0.8358±0.0421 & 23.84±3.11 & 0.1129±0.0317 & 0.8574±0.0441 & 24.84±3.36 & 0.1096±0.0294\\
            MSG-LDM \cite{lin2026multiscale} & 0.8219±0.0404 &23.09±2.89
            & 0.1133±0.0322 & 0.8497±0.0426  & 24.53±3.28 & 0.1062±0.0291\\
            HTSCGAN(Ours) & \textbf{0.8450±0.0406} & \textbf{24.33±2.91} & \textbf{0.1037±0.0298} & \textbf{0.8699±0.0414} & \textbf{25.37±3.21} & \textbf{0.0894±0.0301} \\
            \midrule
            \multicolumn{5}{c}{T1→T2(BraTS2020)} & \multicolumn{1}{c}{T2→T1(BraTS2020)} \\ 
            \cmidrule(lr){2-4} \cmidrule(lr){5-7}
            SynDiff\cite{ozbey2023unsupervised}  & 0.8514±0.0564 & 24.01±2.66 & 0.0924±0.0268 & 0.8564±0.0489 & 21.61±3.68 & 0.0965±0.0316 \\
            RegistFormer\cite{kim2025improving}  & 0.8467±0.0495 & 23.76±2.87 & 0.1008±0.0322 & 0.8479±0.0471 & 21.42±3.41 & 0.1014±0.0304 \\
            HTSCGAN(Ours) & \textbf{0.8566±0.0517} & \textbf{24.42±2.73} & \textbf{0.0901±0.0253} & \textbf{0.8650±0.0462} & \textbf{23.06±3.03} & \textbf{0.0919±0.0301}\\
            \bottomrule
        \end{tabular}
    }

\end{table*}

\begin{table*}[t]
    \centering
    \caption{This table presents a quantitative comparison of model performance on tumor regions. Statistical significance is confirmed by the Wilcoxon signed-rank test (p < 0.005).}
    \resizebox{1.0\textwidth}{!}{
        \begin{tabular}{c ccc ccc}
            \toprule
            \multirow{2}{*}{Method} & \multicolumn{3}{c}{BraTS2020} & \multicolumn{3}{c}{UCSF} \\ 
            \cmidrule(lr){2-4} \cmidrule(lr){5-7}
            & SSIM↑ & PSNR↑ & LPIPS↓ & SSIM↑ & PSNR↑ & LPIPS↓ \\ 
            \midrule
            MILR \cite{chartsias2017multimodal} & 0.6178±0.1056 & 18.89±3.22 & 0.1983±0.0526 & 0.6587±0.1186 & 19.98±3.21 & 0.1351±0.1204 \\
            TSF \cite{han2023explainable} & 0.6224±0.1134 & 19.02±3.11 & 0.1736±0.0623 & 0.6761±0.1237 & 20.77±3.34 & 0.1201±0.1094  \\
            MMGAN \cite{sharma2019missing} & 0.6255±0.1108 & 19.25±3.28 & 0.1871±0.0648 & 0.6828±0.1213 & 21.03±3.44 & 0.1227±0.1113 \\ 
            MMTransformer \cite{liu2023one} & 0.6189±0.1089 & 19.18±3.09 & 0.1902±0.0538 & 0.6704±0.1221 & 20.65±3.38 & 0.1284±0.1147\\
            MSG-LDM \cite{lin2026multiscale} & 0.6102±0.1021 &18.87±3.08
            & 0.1856±0.0617 & 0.6683±0.1202  & 20.69±3.34 & 0.1198±0.1165\\
            HTSCGAN(Ours) & \textbf{0.6356±0.1034} & \textbf{19.76±3.57} & \textbf{0.1653±0.0642} & \textbf{0.7002±0.1209} & \textbf{21.67±3.38} & \textbf{0.1141±0.1072} \\
            \midrule
            \multicolumn{5}{c}{T1→T2(BraTS2020)} & \multicolumn{1}{c}{T2→T1(BraTS2020)} \\ 
            \cmidrule(lr){2-4} \cmidrule(lr){5-7}
            SynDiff \cite{ozbey2023unsupervised} & 0.6279±0.1103 & 19.02±3.41 & 0.1678±0.0682 & 0.6548±0.1121 & 19.64±3.66 & 0.1721±0.703 \\
            RegistFormer \cite{kim2025improving} & 0.6243±0.1094 & 18.76±3.38 & 0.1702±0.0688 & 0.6517±0.1135 & 19.26±3.78 & 0.1779±0.714 \\ 
            HTSCGAN(Ours) & \textbf{0.6335±0.1100} & \textbf{19.25±3.32} & \textbf{0.1548±0.0665} & \textbf{0.6640±0.0938} & \textbf{20.01±3.68} & \textbf{0.1674±0.0695}\\
            \bottomrule
        \end{tabular}
    }

\end{table*}

\begin{figure*}[t]
  \centering
  \includegraphics[width=1.0\linewidth]{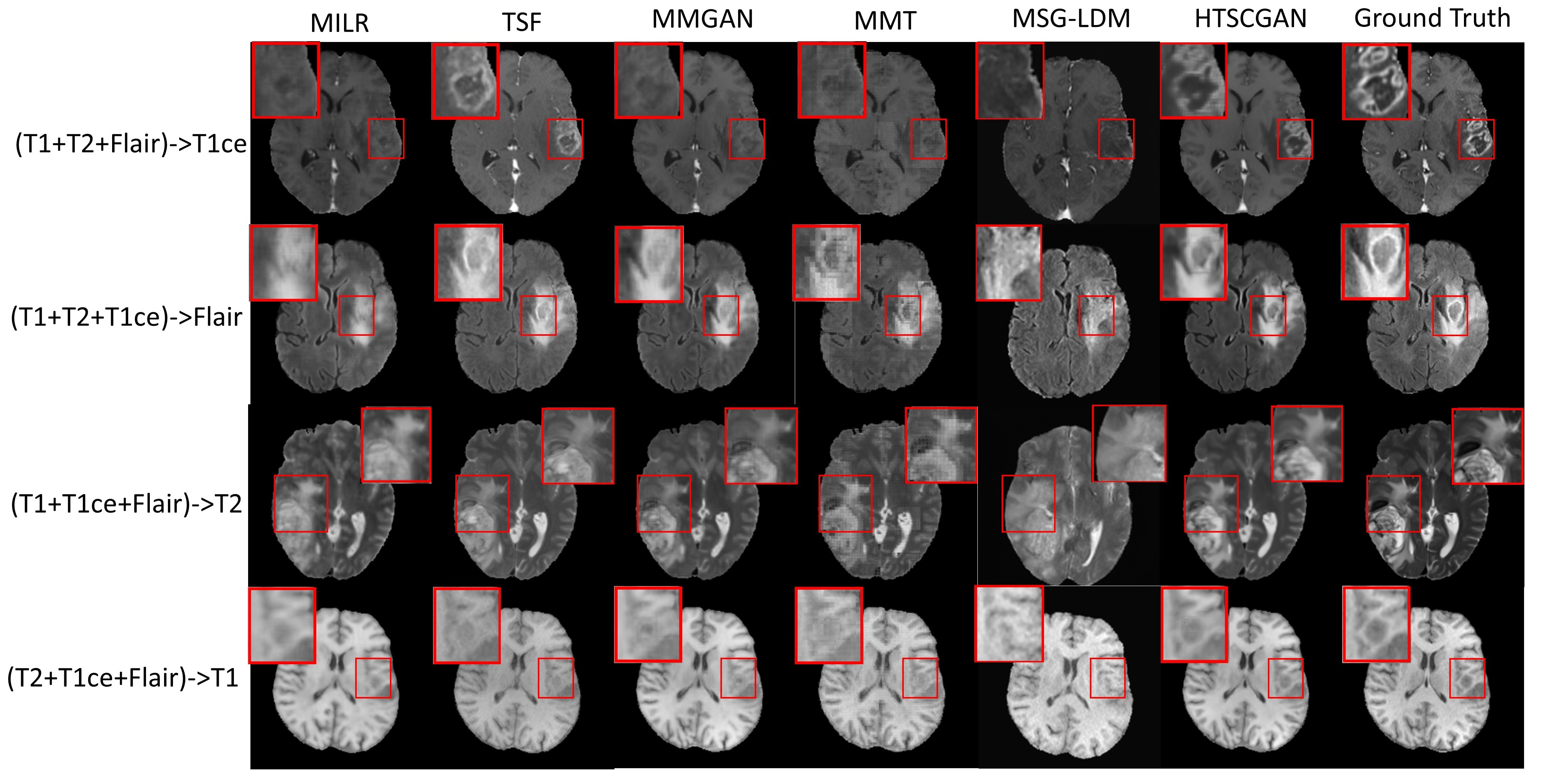}
   \caption{Comparison of translated brain images on BraTS2020 dataset.}
   \label{fig:3}
\end{figure*}

\section{Experiments and Results}

To validate the effectiveness of our model, we compared it with state-of-the-art methods, including MILR \cite{chartsias2017multimodal}, MMGAN \cite{sharma2019missing}, TSF \cite{han2023explainable}, MMTransformer \cite{liu2023one}, SynDiff \cite{ozbey2023unsupervised}, RegistFormer \cite{kim2025improving} and LDM \cite{lin2026multiscale}. MILR is a classic unified brain image translation model, while MMGAN employs a GAN-based architecture. TSF takes a different approach by treating image translation as a sequence-to-sequence prediction task. MMTransformer utilizes a Transformer-based architecture for medical image translation. SynDiff is a one-to-one generative model that employs diffusion models. RegistFormer takes into account the structural information within the pelvic region. MSG-LDM is a unified structure-guided latent diffusion model designed for brain image translation tasks.

\subsection{Comparison of Translation Results}
The comparison of medical image translation performance is shown in Table 1. The results presented in the table represent the average performance across all 14 missing modality cases. We use structural similarity index (SSIM), peak signal-to-noise ratio (PSNR), and LPIPS \cite{zhang2018unreasonable} metrics to evaluate the quality of the generated images. Our model performs better than the other models on both datasets. 

Table 1 also presents a performance comparison with the diffusion-based model SynDiff, as well as RegistFormer, both of which support only single-modality (one-to-one) translation. Accordingly, we specifically evaluated two translation scenarios: T1-to-T2 and T2-to-T1. These were listed separately for comparison. Our model demonstrated a clear advantage in terms of translation quality in both cases, further highlighting its superior performance across different modalities.

Notably, compared with diffusion-based approaches such as MSG-LDM, our method exhibits clear advantages under limited data conditions. Specifically, in our experimental setup, only 20 tumor-containing slices from a single patient are used for training. Under such data-scarce conditions, LDM-based methods tend to suffer from degraded generation quality due to insufficient data for stable diffusion training. In contrast, our approach maintains high-quality image synthesis, highlighting its stronger generalization ability and data efficiency.

Additionally, we conducted a separate comparison of the metrics specifically for the tumor regions. Specifically, for each image, we identified the smallest enclosing rectangular region that fully covers the tumor area, and retained only those regions with a spatial size greater than 32×32 pixels for evaluation. The detailed data is presented in Table 2. From Table 2, it is evident that our model exhibits superior performance in terms of metrics related to the tumor regions compared to other models. The other models did not account for the structural information of the tumor regions in brain images, resulting in suboptimal generation outcomes in these areas.

Figure 3 offers a visual comparison of the generated images from different models across modalities, demonstrating that our model generates more complete and well-defined tumor structures. This is particularly evident in the T1ce modality, where our model exhibits clearer tumor regions compared to other approaches.

\begin{table*}[t]
    \centering
    \caption{Comparison of average dice scores for segmentation results on generated full sequences between HTSCGAN and other models on BraTS2020 dataset. The results are statistically significant, validated by the Wilcoxon signed-rank test (p < 0.005).}\label{tab:2} 
    \resizebox{0.7\textwidth}{!}{
    \begin{tabular}{ccccc}
        \toprule
        Method   & ET  & TC & WT &Average \\ \midrule
        W/O Translation & 0.653±0.303     & 0.755±0.208   & 0.816±0.326
    & 0.741±0.279    \\
        MILR \cite{chartsias2017multimodal} & 0.671±0.306    & 0.762±0.212   & 0.820±0.331 & 0.751±0.283 \\
        TSF \cite{han2023explainable} & 0.636±0.301     & 0.771±0.222   & 0.832±0.287    & 0.746±0.270 \\
        MMGAN \cite{sharma2019missing} & 0.660±0.314     & 0.769±0.205   & 0.814±0.328    & 0.748±0.283    \\
        MMT \cite{liu2023one} & 0.671±0.305     & 0.771±0.217   & 0.822±0.331    & 0.755±0.284    \\
        MSG-LDM \cite{lin2026multiscale} & 0.662±0.308     & 0.766±0.214   & 0.815±0.329    & 0.748±0.284    \\
        HTSCGAN(Ours) & \textbf{0.677±0.297}    & \textbf{0.781±0.205}   & \textbf{0.836±0.336}  & \textbf{0.765±0.279}\\
        \bottomrule
    \end{tabular}
    }

\end{table*}

\begin{table}[t]
    \centering  
    \caption{Ablation study on image quality and segmentation performance on BraTS2020 dataset.}
    \label{tab:3} 
    \resizebox{1.0\columnwidth}{!}{
    \begin{tabular}{ccccccccc}
        \toprule
        Method & \multicolumn{3}{c}{Translation Metric} & \multicolumn{4}{c}{Segmentation Metric} \\ \cmidrule(lr){2-4} \cmidrule(lr){5-8}
               & SSIM  & PSNR & LPIPS & ET  & TC & WT  & Average \\ \midrule
        W/O all & 0.8314\footnotesize{±0.0407}     & 23.32\footnotesize{±3.06}   & 0.1080\footnotesize{±0.0308}   & 0.655±0.311     & 0.756±0.217   & 0.822±0.292  & 0.744±0.273 \\
        W/O PCM & 0.8397\footnotesize{±0.0402}     & 23.78\footnotesize{±2.96}   & 0.1186\footnotesize{±0.0298}   & 0.659±0.315     & 0.763±0.223   & 0.831±0.284  & 0.751±0.274 \\
        W/O PC  & 0.8469\footnotesize{±0.0406}      & 23.99\footnotesize{±3.02}   & 0.1096\footnotesize{±0.0303}   & 0.658±0.318    & 0.759±0.221   & 0.832±0.319  & 0.750±0.286 \\
        W/O SAE & 0.8429\footnotesize{±0.0408}     & 23.84\footnotesize{±2.99}   & 0.1130\footnotesize{±0.0311}   & 0.668±0.313     & 0.768±0.226   & 0.831±0.297  & 0.756±0.279 \\
        HTSCGAN & \textbf{0.8502\footnotesize{±0.0401}}    & \textbf{24.18\footnotesize{±2.96}}   & \textbf{0.0985\footnotesize{±0.0299}}   & \textbf{0.677±0.297}    & \textbf{0.781±0.205}   & \textbf{0.836±0.336} & \textbf{0.765±0.279} \\
        \bottomrule
    \end{tabular}
    }

\end{table}

\subsection{Downstream Segmentation}
To demonstrate the applicability of our generated images in clinical diagnosis, we further utilized them in a tumor segmentation task. Specifically, we employed a U-Net \cite{ronneberger2015u} as the segmentation network, randomly replacing modalities with our generated images for segmentation. Segmentation targets included the tumor core (TC), enhancing tumor (ET), and whole tumor (WT). Dice scores between the masks generated by the model and the ground truth masks for these three regions are reported in Table 3. For the condition labeled "W/O Translation," we utilize a zero-filled image for padding, while the remaining instances are filled with generated images. The values reported in the table are averaged over all missing modality scenarios. As shown, all models exhibit improved performance when the missing images are filled in, compared to scenarios where the missing data remains unaddressed. Notably, our model achieved the best segmentation performance among all models, suggesting that the high fidelity and clinical significance of the images generated by HTSCGAN.

\begin{figure*}[t]
  \centering
  \includegraphics[width=0.7\linewidth]{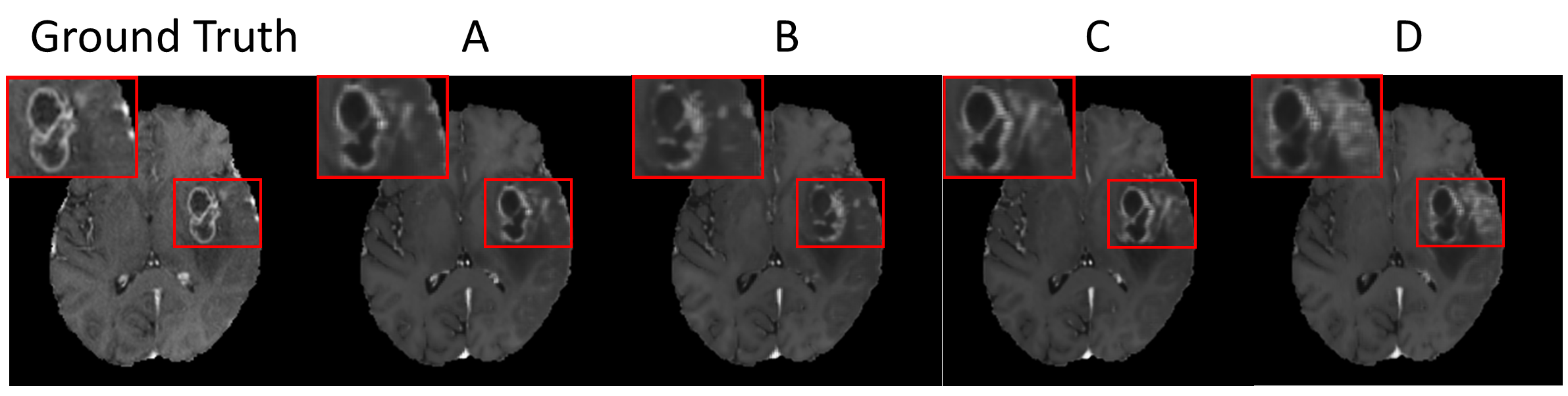}
   \caption{Comparative brain image outputs with ablation variants on BraTS2020 dataset. (A) the output of HTSCGAN; (B) the output without PCM; (C) the output without PC; (D) the output without SAE.}
   \label{fig:4}
\end{figure*}

\subsection{Ablation Study}
\textbf{Patch Contrastive Module.} From the comparative data in Table 4, it is evident that the inclusion of this module improves performance across all three image quality metrics as well as segmentation accuracy. This demonstrates that the module helps the generator produce higher-quality brain images. As shown in Figure 4 B, the absence of the contrastive module leads to unclear tumor structures and blurriness. 

\noindent\textbf{Patch-Classifier.} The data presented in Table 4 indicate that the absence of the PC module leads to a degradation in image performance. Moreover, segmentation experiments reveal that the PC has a positive impact on the downstream segmentation task, suggesting that the tumor regions identified by the PC are closer to the real tumor regions.

\noindent\textbf{Structure-Aware Encoder.} As shown in Table 4, the model equipped with the SAE achieves better image quality.  In Figure 4 D, a white area appear around the tumor region in the T1ce modality, which do not align with the actual anatomical structure. This indicates that the SAE effectively enhances the model’s ability to produce more meaningful tumor representations.

\section{Conclusion}

This paper proposes HTSCGAN, a novel brain image translation model that incorporates tumor region structural information through three key components: PCM, PC, and SAE. Our approach enables the model to generate more realistic and clinically meaningful brain images with well-defined tumor regions. Furthermore, the generated images demonstrate superior performance in the downstream tumor segmentation tasks, underscoring the model's effectiveness in enhancing both image quality and segmentation accuracy.

{
    \bibliographystyle{splncs04}
    \bibliography{main}
}
\end{document}